\documentclass[11pt]{article}

\usepackage[preprint]{acl}

\usepackage{times}
\usepackage{latexsym}

\usepackage[T1]{fontenc}

\usepackage[utf8]{inputenc}

\usepackage{microtype}

\usepackage{inconsolata}

\usepackage{graphicx}

\usepackage{listings}
\usepackage[ruled,vlined]{algorithm2e}
\usepackage{multirow}
\usepackage{tabularx}
\usepackage{xcolor}     
\usepackage{tcolorbox}
\tcbuselibrary{breakable}
\usepackage{booktabs}   
\usepackage{pifont}     
\usepackage{makecell}   

\usepackage{amsmath}
\usepackage{amssymb}

\definecolor{policy}{HTML}{aee2cb}
\definecolor{actor}{HTML}{faca7f}
\definecolor{critic}{HTML}{fa99a9}
\usepackage{pdfpages}  

\title{SciCurateBench: Curating Experimental Data from Heterogeneous Scientific Literature with Verbal-Action Agents}

\author{Minuk Ma, Kaiyun Guo, Jiarui Ding \\
  University of British Columbia \\
  \texttt{\{minukma, gky0722\}@cs.ubc.ca}, \texttt{jiarui.ding@ubc.ca}}

\begin{document}
\maketitle
\begin{abstract}
Scientific research increasingly depends on curating experimental data from literature. Scientific data curation requires verifying exact experimental conditions (e.g., specific assays, clinical outcomes, and data availability) across long, heterogeneous files, which is a highly time-consuming process. Existing benchmarks evaluate only broad relevance on short queries, failing to capture this complexity. We introduce SciCurateBench, a benchmark of 200 expert-annotated papers from major life-science journals that challenges systems to verify conjunctive biological experimental criteria from raw, unprocessed scientific literature. We show that plain LLMs and popular reasoning, tool-using, and feedback-based agents all struggle with this task. To study what makes the task tractable, we further propose LangPoli, an integrative system in which a policy agent generates natural language action guidelines that actor agents convert into executable tools. LangPoli showed strong performance on both Paper Classification (83.16\% F1 vs.\ 58.22\% for its base LLM) and Structured Information Retrieval, and also improves re-ranking on the public benchmarks of LitSearch, BEIR, and TREC. We further explore curriculum transfer, a method that distills generalizable strategies from solved tasks to accelerate the solving of related problems. We hope this work facilitates research in truly practical scientific AI assistance. The anonymized pre-release can be found from \url{https://anonymous.4open.science/r/scicuratebench-D885/}. 
\end{abstract}

\section{Introduction}

Modern scientific research increasingly begins with dataset curation, systematically gathering experimental data from published literature that satisfy precise criteria. This is especially common in life sciences research to build atlases by merging multiple similar experimental data \cite{atlas_1,atlas_3,atlas_2,scatlasvae,scbasecount}. For example, researchers may seek "all publicly available single-cell RNA sequencing (scRNA-seq) data from cancer patients with associated immunotherapy response outcomes". This seemingly straightforward task is extraordinarily time-consuming in practice. Scientific papers are long, with extensive supplementary materials, and critical information may be embedded in externally hosted data (e.g., GEO \cite{geo} or Mendeley Data \cite{mendeleydata}). Researchers must verify multiple conditions simultaneously: the correct assay type (scRNA-seq, not bulk RNA-seq), disease context (cancer), clinical information (treatment outcomes), and crucially, that data are newly generated (not reanalyzed from prior work) and publicly available. A single paper might report multiple experiments, and only one of them may meet all criteria.

While modern search engines allow for the identification of potentially relevant papers, the challenge lies in verifying whether each paper precisely satisfies all the conditions in the query. This precision requirement, combined with the length and heterogeneity of scientific documents, makes automated curation still far from being satisfactorily solved.

Recent advances in large language models (LLMs) \cite{gpt3,gpt4_technical_report,claude3,palm} have shown promise for literature search tasks. Powerful models such as GPT-4o \cite{gpt4_technical_report} have demonstrated effectiveness in re-ranking candidate documents \cite{rankgpt} and identifying citations for specific claims \cite{citeme}. 
However, existing literature search benchmarks focus on broad relevance ranking or single-claim attribution rather than confirming that papers precisely meet multiple conjunctive experimental criteria. Also, most of them do not include supplementary files or hosted data as search targets, deviating from practical needs. 

To address this gap, we introduce SciCurateBench, a benchmark comprising 200 expert-annotated papers from mainstream scientific journals, with the goal of facilitating research in realistic scientific data curation. The benchmark includes two tasks that mirror real-world workflows: Science Paper Classification, which requires determining whether experiments meet specified criteria, and Structured Information Retrieval, which involves extracting experimental metadata from papers that satisfy the search conditions. 

We show that popular LLM reasoning methods achieve only suboptimal performance on the benchmark. To further push performance and study the factors contributing to success, we propose LangPoli, a multi-agent framework that adaptively creates tools to retrieve information from heterogeneous documents and verify precise conditions in the query. Unlike code-action agents that couple planning and implementation in a single generation step, LangPoli separates a natural-language policy from environment-specific code synthesis. This enables the creation of dynamic tools tailored to each document's unique structure. Actor agents then convert these verbal action guidelines into executable code to collect observations. 
A critic agent evaluates the accumulated evidence against the query predicates and provides info-gap feedback to the policy agent, creating a closed loop that enables self-improvement over iterations. 

We note that concepts such as the plan/action structure \cite{ts_llm,plan_and_act,agent_oriented}, code as action instead of predefined sets \cite{codeact,craft,toolmaker,dynasaur}, evaluation using LLM agents \cite{reflexion,self_check,agent_as_a_judge,reinforce_llm}, and adaptive tool usage \cite{toolformer,adaptive_tool_use,tool_gen,evolve_tools} have been established in the field, and LangPoli is a domain-specific adaptation of these components for the underexplored task of scientific data curation. Still, benchmarking LangPoli and other methods on SciCurateBench reveals the difficulty of the task and provides insights into the factors contributing to success. 

\begin{figure*}[t] 
    \centering
    \includegraphics[width=\textwidth]{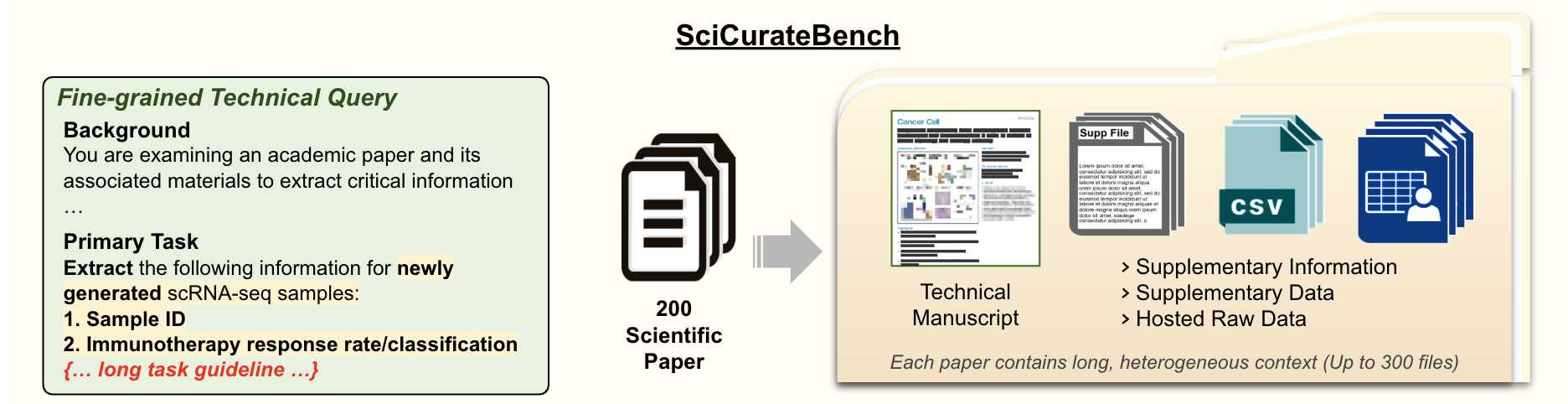}
    \caption{SciCurateBench promotes adaptive tool creation for efficiently exploring, analyzing, and summarizing heterogeneous scientific files, including manuscripts, supplementary data, and raw datasets.}
    \label{fig:scicuratebench} 
\end{figure*}

We provide comprehensive benchmarking of popular agentic AI baselines on scientific data curation, including plain LLMs, reasoning agents, tool-using agents, tool-creating agents, and feedback-based agents. Our experiments show that LangPoli is the only method that performs consistently well on both tasks of SciCurateBench, reaching 83.16\% F1 on paper classification (vs.\ 58.22\% for its base LLM) and 69.43\% F1 on structured information retrieval (vs.\ 59.13\% for the strongest baseline). The framework also shows strong performance on public benchmarks of LitSearch \cite{litsearch}, BEIR \cite{beir}, and TREC \cite{trec}, particularly excelling on long documents where its created tools enable efficient targeted exploration. We show that the tools generated by LangPoli maintain high execution success rates and code maintainability scores while effectively supporting the exploration process.

Finally, we explore curriculum transfer, a form of knowledge distillation where successful tool creation sequences are refactored and transferred as structured curricula to accelerate performance on related tasks. This approach enables the system to leverage problem-solving strategies learned from previous curation tasks, further improving efficiency and effectiveness on new scientific data curation problems.

\section{Related Work}
\textbf{Scientific literature search and curation.}
The challenge of finding and processing scientific literature has driven the development of research assistance systems. Classical retrieval models such as BM25 \cite{bm25}, a term-based ranking method derived from the probabilistic relevance framework, remain a strong baseline. Information-retrieval benchmarks such as BEIR \cite{beir}, TREC \cite{trec}, and MTEB \cite{mteb} have facilitated research on literature search by evaluating across various domains and query types. As more targeted research, LitSearch \cite{litsearch} simulates literature search from machine learning papers. However, these benchmarks primarily assess candidate ranking by broad relevance to user queries. 

Recent papers applied LLMs to retrieve relevant documents or data, and standardize \cite{pasa,bright,curie,mole,openscholar}. CiteME \cite{citeme} evaluates whether LLM agents can identify which paper a text excerpt references, targeting citation attribution for fact-checking. While RankGPT \cite{rankgpt} and CiteME have benchmarked GPT-4o and Claude 3.5 on re-ranking and citation tasks, evaluation of other agentic AI families remains unexplored. An extended discussion is provided in Appendix~\ref{app:related_work}. In contrast to previous works, our SciCurateBench addresses a different yet practical bottleneck of experimental data curation. While researchers can gather top candidates using search engines, verifying whether each candidate actually published the desired experimental data (e.g., "published data using technique X with metric Y and metadata Z") poses additional layers of grounding, verification, and cross-checking challenges. 

\textbf{LLM reasoning and tool creation.}
Recent advances have enhanced LLM reasoning through structured prompting \cite{cot}, tool integration \cite{latm,codeact,craft,toolmaker}, and feedback-based refinement \cite{reflexion,agent_as_a_judge,reinforce_llm}. LangPoli is structurally close to these methods but decouples policy (guidelines) from environment-specific implementation (code). Retrieval-augmented LLMs \cite{speculative_rag,long_context_rag} ground claims with evidence from web-based \cite{search_o1,web_thinker} or static corpus \cite{hippo_rag2,openscholar}. As the field is rapidly evolving, we point the readers to a recent survey \cite{rag_survey} for more related methods. In this work, we focus our scope on scientific data curation and select baselines that are widely acknowledged, rather than extensive lists of agentic coders.

\textbf{Scientific AI agents.}
Recent works have begun applying LLM agents to scientific workflows \cite{scienceagentbench,aiscientist,bixbench,chemcrow,aiscientist_nature,co_scientist,cellagent}. Existing research on scientific AI agents predominantly targets downstream tasks of hypothesis generation, experimental design, data analysis, or model fitting, and assumes that curated input data is already available. By contrast, the upstream bottleneck of curating experimental data from raw, heterogeneous primary literature remains underexplored. SciCurateBench and LangPoli target this upstream gap, and we expect strong synergy with existing research on scientific AI agent.

\section{SciCurateBench Dataset}
We introduce \textbf{SciCurateBench}, a benchmark designed to evaluate AI systems on realistic data curation tasks. 

\subsection{Dataset Construction}
\textbf{Paper Selection.} As shown in Figure \ref{fig:scicuratebench}, we curated 200 papers from major life science journals (\textit{Nature}, \textit{Cell}, \textit{Science}, etc.; 2016--2025) focusing on two query types reflecting realistic workflows: (1) scRNA-seq with immunotherapy response: identifying newly generated single-cell RNA sequencing data from cancer patients with associated clinical outcomes, and (2) paired scRNA-seq and scTCR-seq: identifying papers with simultaneously profiled transcriptomic and T-cell receptor sequences across disease conditions. Papers were identified through PubMed searches using domain-specific keywords (``single-cell RNA sequencing,'' ``scTCR-seq,'' ``immunotherapy response,'' ``checkpoint blockade'') and expert-curated paper lists from atlas construction projects~\cite{scatlasvae}. 

\textbf{Annotation Protocol.} Three domain experts with a life science background (3--12 years experience) performed annotation by examining full manuscripts, supplementary materials, and hosted data. Annotations were conducted in two stages: primary annotation (160 hours) followed by independent verification with disagreement resolution (additional 40 hours). The resulting two annotations yielded a Cohen's $\kappa$ of 0.932. The detailed protocol is described in Appendix~\ref{app:annotation}. 

\textbf{Dataset Characteristics.} For Query 1 (scRNA-seq + immunotherapy), the dataset contains: 38 papers with newly generated data meeting all criteria (Class 3), 12 papers reanalyzing existing data (Class 2), and 50 papers that fail to meet all conditions (Class 1). For Query 2 (paired scRNA-seq + scTCR-seq), papers are distributed across 8 disease categories (Classes 1--8). The per-class distribution and summary statistics can be found in Appendix~\ref{app:characteristics} (Figure~\ref{fig:dataset_stats}). This distribution reflects realistic prevalence: newly generated data meeting stringent criteria are relatively rare, requiring substantial exploration.

SciCurateBench papers average 24,318 tokens (median 21,792, IQR 15,748--31,050, range: 931--70,121), with 95\% exceeding 8K tokens. Each paper comprises: (1) main manuscript (PDF parsed to text), (2) supplementary files in multiple formats: XLSX tables, CSV metadata, plain text, and (3) raw data (.mtx, .Rds, etc.). 
We mostly preserve raw formats to test whether agents can dynamically adapt to diverse documents, moving away from idealized, pre-curated formats that many benchmarks assume. 

\textbf{Dataset Setup.}
The release format and preprocessing steps are described in Appendix~\ref{app:setup}. 

\subsection{Task Definitions}
\textbf{Task 1: Science Paper Classification.} Given a paper and classification guideline, determine the class label $y \in \mathcal{Y}$ where $\mathcal{Y} = \{1, 2, 3\}$ for Query 1 or $\mathcal{Y} = \{1, 2, 3, 4, 5, 6, 7, 8\}$ for Query 2. The guideline specifies conjunctive predicates that must be verified: assay type $\land$ data novelty $\land$ subject population $\land$ outcome measurement $\land$ data accessibility. For Query 1, agents must distinguish newly generated scRNA-seq data with immunotherapy response (Class 3) from reanalyzed existing data (Class 2) or irrelevant papers (Class 1). 
Agents receive basic file operation tools (\texttt{read\_file}, \texttt{search\_keywords}, \texttt{extract\_citations}, etc.) but must autonomously determine exploration strategies. Accuracy and F1 score are used as primary metrics. 

\textbf{Task 2: Structured Information Retrieval.} For papers classified as Class 3 in Query 1, extract structured tuples $(s_i, r_i)$ where $s_i$ is a sample/patient identifier and $r_i$ is the immunotherapy response outcome. 
% ...
Valid response classifications include: Response Evaluation Criteria in Solid Tumors (RECIST) criteria (CR, PR, SD, PD), binary labels (Responder/Non-responder), pathological responses, or survival metrics explicitly linked to immunotherapy outcomes. This task requires cross-referencing information: sample IDs may appear in one table while response data appears in another, necessitating joins across files. Agents must create custom parsing tools adapted to each paper's unique table structure and field naming conventions. Evaluation metrics are based on Precision, Recall, and F1 for exact $(s_i, r_i)$ matching, partial credit for correct $s_i$ but incorrect $r_i$ or vice versa. 

\textbf{Why So Difficult?}
In Task 1, a representative challenge arises when a paper contains some samples with the correct assay type but no associated outcome measurement, while other samples report the outcome but use a slightly different assay. The required predicates must therefore be satisfied jointly \textit{at the sample level}, not merely present somewhere in the paper. Beyond addressing the practical need to curate experimental metadata, Task 2 complements the classification task in Task 1: it grounds Class 3 membership in the data itself. Many papers that look like Class 3 do not publish their full metadata, such as per-sample response labels (Appendix~\ref{app:difficulty}). Researchers cannot know what a paper truly contains until they inspect every supplementary file and its raw data. 




\begin{figure*}[t]
    \centering
    \includegraphics[width=0.95\textwidth]{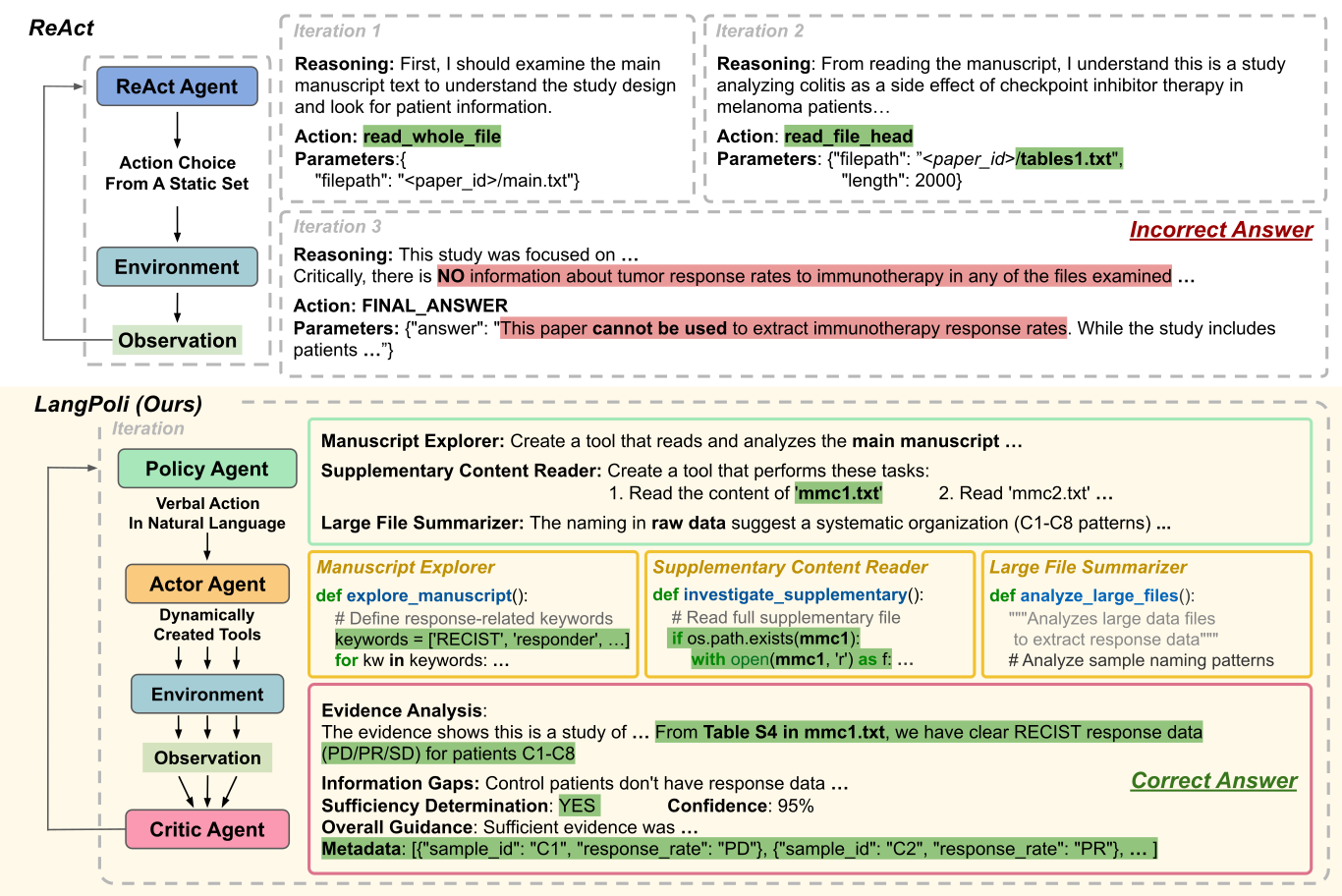}
    \caption{Overview of LangPoli on SciCurateBench. LangPoli creates an adaptive tool based on the verbal action to interact with the environment, while ReAct chooses an action from a fixed set. }
    \label{fig:langpoli_overview}
\end{figure*}

\section{LangPoli Framework}
\label{sec:langpoli}

\subsection{Problem Formulation and Motivation}

Scientific data curation can be formalized as follows. Given a paper environment $E$, comprising heterogeneous files, and a task objective $G$, specifying conjunctive predicates to verify, the goal is to produce a judgment $J$, supported by an evidence collection $V$. Traditional agent frameworks assume a fixed action space with $k$ actions: $\mathcal{A} = \{a_1, a_2, \ldots, a_k\}$, where each action $a_i$ has predefined semantics (e.g., \texttt{read\_file(path)}, \texttt{search\_keyword(query)}). 

Compared to this, LangPoli features \textbf{verbal action generation}: rather than selecting $a_t \in \mathcal{A}$, the framework generates natural language action descriptions $\tilde{a}_t$ that actor agents translate into paper-specific executable code $c_t$, effectively creating a dynamic action space $\mathcal{A}_E$ adapted to the environment $E$. Figure~\ref{fig:langpoli_overview} illustrates this distinction, where ReAct iteratively applies fixed actions and may miss crucial connections between document sections, whereas LangPoli synthesizes custom tools to navigate each paper's unique structure.

\subsection{Framework Architecture}

LangPoli comprises three specialized agents in an iterative refinement loop:

\textbf{Policy Agent $\pi$:} At iteration $t$, given accumulated evidence $V_t$, exploration history $H_t = \{(\tilde{a}_1, \Delta V_1, r_1), \ldots, (\tilde{a}_{t-1}, \Delta V_{t-1}, r_{t-1})\}$, and task objective $G$, the policy agent generates the verbal action:

\begin{equation}
\tilde{a}_t = \pi(V_t, H_t, G; \theta_\pi)
\end{equation}

\noindent where $\theta_\pi$ parameterizes the policy LLM. Unlike discrete action selection, $\tilde{a}_t$ is a natural language description specifying \textit{what} to explore. 
An example is shown in Appendix~\ref{app:verbal_action}. 
This verbal action provides strategic guidance while delegating implementation details to actor agents that understand the specific environment structure.

\textbf{Actor Agents $\{\mu_1, \ldots, \mu_K\}$:} Given verbal action $\tilde{a}_t$ and environment context $\mathcal{C}_E$, actor agent $\mu_k$ generates executable code:

\begin{equation}
c_t^{(k)} = \mu_k(\tilde{a}_t, \mathcal{C}_E; \theta_{\mu_k})
\end{equation}

The generated code $c_t^{(k)}$ consists of Python functions tailored to the current paper's structure. Execution produces observations:

\begin{equation}
\Delta V_t^{(k)} = \text{execute}(c_t^{(k)}, E)
\end{equation}

For complex tasks, we employ \textbf{Mixture-of-Collaborative-Actors (MoCA)}: multiple specialized actors $\{\mu_1, \ldots, \mu_K\}$ operate in parallel, each focusing on different evidence-gathering strategies. For instance, in Structured Information Retrieval: $\mu_1$ explores the manuscript text, $\mu_2$ focuses on supplementary files, and $\mu_3$ summarizes large hosted data. Observations are aggregated: $\Delta V_t = \bigcup_{k=1}^K \Delta V_t^{(k)}$, forming a comprehensive evidence collection $V_{t+1} = V_t \cup \Delta V_t$.

\begin{table*}[htbp]
  \centering
  \caption{Benchmarking performance on Science Paper Classification. Std for F1 score was provided. The metrics for Class 3 were additionally provided. LangPoli (full) achieves the best accuracy, F1, and Class 3 F1.}
  \label{tab:paper_classification}
  \resizebox{0.85\textwidth}{!}{
  \begin{tabular}{l|l|ccccccc}
    \toprule
    \textbf{Group} & \textbf{Method} & \textbf{Acc (\%)} & \textbf{Prec (\%)} & \textbf{Rec (\%)} & \textbf{F1 (\%)} & \textbf{C3 Prec (\%)} & \textbf{C3 Rec (\%)} & \textbf{C3 F1 (\%)} \\
    \midrule
    \multirow{2}{*}{Plain LLM} 
    & gpt-4o & 65.33 & 61.08 & 64.99 & 61.00 (2.46) & 67.38 & 83.33 & 74.46 \\
    & claude-3.5 & 67.67 & 63.48 & 57.76 & 58.22 (2.23) & \textbf{91.67} & 52.63 & 66.59 \\
    \midrule
    \multirow{2}{*}{Reasoning}
    & CoT         & 75.00 & 71.16 & 73.13 & 71.96 (1.97) & 77.39 & 78.06 & 77.71 \\
    & Self-Con    & 73.33 & 68.56 & 71.38 & 69.67 (0.61) & 76.96 & 78.94 & 77.91 \\
    \midrule
    \multirow{4}{*}{Tool}
    & ReAct       & 79.67 & 75.68 & 79.05 & 76.95 (1.01) & 79.16 & 86.84 & 82.81 \\
    & ChatCoT     & 76.59 & 71.26 & 74.94 & 72.21 (0.76) & 77.08 & 87.72 & 82.02 \\
    & LATM        & 77.33 & 73.18 & 76.30 & 74.32 (4.57) & 75.52 & 80.69 & 77.99 \\
    \midrule
    \multirow{2}{*}{Feedback}
    & Reflexion   & 82.48 & 80.71 & 77.63 & 78.74 (0.81) & 79.10 & 89.47 & 83.94 \\
    & Self-Refine & 81.47 & 78.20 & 76.86 & 77.29 (1.08) & \textbf{85.16} & 78.94 & 81.84 \\
    \midrule
    \multirow{2}{*}{LangPoli}
    & LangPoli-Basic    & \textbf{83.00} & \textbf{80.72} & \textbf{79.65} & \textbf{79.76} (3.65) & 78.58 & \textbf{92.98} & \textbf{85.15} \\
    & LangPoli (full)   & \textbf{86.33} & \textbf{82.74} & \textbf{83.92} & \textbf{83.16} (0.94) & 84.70 & \textbf{92.10} & \textbf{88.19} \\
    \bottomrule
  \end{tabular}%
  }
\end{table*}

\textbf{Critic Agent:} Given accumulated evidence $V_{t+1}$ and task objective $G$, the critic produces structured feedback $r_t = \text{critic}(V_{t+1}, G; \theta_c)$, which consists of an evidence summary (key discoveries), info-gaps ($\{p_i \in G : p_i \text{ not verified}\}$), a sufficiency indicator, a confidence score $\rho_t \in [0, 1]$, and next-step guidance for $\pi$ (full schema in Appendix~\ref{app:critic}). 
The sufficiency determination acts as a termination condition: when $\text{sufficiency} = \text{True}$ and $\rho_t > \tau$ (confidence threshold), exploration concludes. Otherwise, identified information gaps guide the policy agent's next verbal action $\tilde{a}_{t+1}$.

\section{Experiments}

\subsection{Experimental Setup}
\textbf{Models and Baselines.} We evaluate multiple approaches: base LLMs (GPT-4o \cite{gpt4_technical_report}, Claude-3, Claude-3.5, Claude-3.7 \cite{claude3}), reasoning methods (CoT \cite{cot}, Self-Consistency \cite{self_con}), tool-based agents (ReAct \cite{react}, ChatCoT \cite{chatcot}, LATM \cite{latm}), and feedback approaches (Reflexion \cite{reflexion}, Self-Refine \cite{self_refine}). All models are adapted to SciCurateBench. We implement two LangPoli variants: LangPoli-Basic (single actor) and LangPoli (full, MoCA). Model versions, the smaller LLMs evaluated on the public benchmarks, and the hardware are listed in Appendix~\ref{app:exp_details}. 

\textbf{Implementation Details.} The implementations use Claude-3.5 as the base LLM unless otherwise noted. Tool creation is restricted to Python and token usage is capped at 30,000 tokens per environment. All experiments run in zero-shot without additional training. 

\textbf{Evaluation Metrics.} Paper Classification uses accuracy and F1 score, with separate reporting for Class 3 (C3), which is the class that meets all criteria in the query. In real-world scenarios, Class 3 Recall is important because even a single additional paper meeting the criteria may contain a large amount of patient data, contributing significantly to the desired data atlas. Structured Information Retrieval measures precision, recall, and accuracy for sample (patient) ID, clinical response, and their pairs. We also analyze tool effectiveness based on Maintainability Index (MI) \cite{mi,welker2001software}. For LitSearch, BEIR, and TREC, we follow the metrics used in the original papers: Recall@K and NDCG@K. 

\subsection{Performance on SciCurateBench: Paper Classification}

Table \ref{tab:paper_classification} presents the performance of all methods on the paper classification task. Plain LLMs (GPT-4o and Claude-3.5) achieved F1 scores of only 61.00\% and 58.22\%, respectively. This highlights the difficulty of extracting sparse key information from long technical content. Reasoning approaches showed large improvements: CoT and Self-Con boosted F1 scores by 13.74 and 11.45 points over Claude-3.5-Sonnet, showing that conceptual task decomposition enhances information retrieval.

\begin{table}[t]
  \centering
  \caption{Classification F1 (\%) on the longest-quartile papers (top 25\% by token count, $>$31K tokens) versus the full benchmark. LangPoli degrades far less than Claude-3.5 on long papers. }
  \label{tab:length_robustness}
  \resizebox{0.40\textwidth}{!}{
  \begin{tabular}{lccc}
    \toprule
    \textbf{Method} & \textbf{Full (\%)} & \textbf{Longest 25\% (\%)} & \textbf{$\Delta$} \\
    \midrule
    GPT-4o          & 61.00 & 59.47 & $-1.53$ \\
    Claude-3.5      & 58.22 & 43.83 & $-14.39$ \\
    LangPoli (full) & \textbf{83.16} & \textbf{79.90} & $-3.26$ \\
    \bottomrule
  \end{tabular}%
  }
\end{table}

Tool-based agents consistently outperformed pure reasoning approaches. ReAct achieved a 76.95\% F1 score, showcasing the advantage of information retrieval tools. Although LATM is capable of generating tools, it constructs static, high-level tool templates before interacting with rich environments, effectively reducing tool use to simple keyword-based search. Feedback-based agents (Reflexion and Self-Refine) also surpassed plain LLMs and reasoning-only approaches. LangPoli-Basic achieved a competitive 79.76\% F1 score (21.54 points improvement over the plain LLM baseline), while the full LangPoli framework reached 83.16\% F1 and 88.19\% Class 3 F1, ahead of all other baselines. This shows the importance of dynamic tool creation and the feedback loop for this task. 

We stratify the benchmark by token count and report F1 on the longest quartile ($>$31K tokens) in Table~\ref{tab:length_robustness}. Claude-3.5 drops F1 by 14.4 points, whereas LangPoli degrades F1 by 3.3 points and retains an absolute F1 of 79.90\%, supporting the importance of targeted, tool-mediated exploration. 

\begin{table}[htbp]
  \centering
  \caption{Benchmarking performance on Structured Information Retrieval. F1 measures the completeness of the pairs. Pat. ID measures Acc on sample (patient) ID and Resp. Rate measures Acc on the response rate information. LangPoli (full) achieves the best F1 and misses the fewest patients.}
  \label{tab:structured_info_retrieval}
  \small 
  \resizebox{\columnwidth}{!}{%
\begin{tabular}{l|c|ccc}
    \toprule
    \textbf{Method} & \textbf{F1 (\%)} & \textbf{Pat. ID (\%)} & \textbf{Resp.} & \textbf{Miss.} \\
    & (std) & & \textbf{Rate (\%)} & \textbf{Pat.} ($\downarrow$) \\
    \midrule
    ReAct          & 53.57 (1.19) & 59.73 & 75.63 & 5.93 \\
    ChatCoT        & 59.13 (4.40) & 65.60 & 80.43 & 6.57 \\
    LATM           & 11.50 (4.62) & 20.43 & 29.27 & 12.10 \\
    \midrule
    Reflexion      & 57.33 (8.81) & 62.23 & 74.37 & 5.75 \\
    SelfRefine     & 48.47 (5.56) & 52.47 & 60.70 & 8.02 \\
    \midrule
    LangPoli-Basic       & \textbf{62.97} (3.55) & \textbf{74.77} & \textbf{82.57} & \textbf{3.81} \\
    LangPoli (full)      & \textbf{69.43} (5.34) & \textbf{81.77} & \textbf{90.30} & \textbf{1.63} \\
    \bottomrule
  \end{tabular}%
}
\end{table}

\subsection{Performance on SciCurateBench: Structured Information Retrieval}
Table \ref{tab:structured_info_retrieval} summarizes performance on Structured Information Retrieval. Plain LLMs are omitted as this task fundamentally requires tool usage. Tool-using agents ReAct and ChatCoT achieved reasonable F1 scores of 53.6\% and 59.1\%, setting performance baselines. LATM struggled due to difficulties in developing master tools for highly heterogeneous paper structures. Feedback-based agents, particularly Reflexion, demonstrated a reasonable F1 score of 57.3\%, underscoring the importance of environmental interaction and self-reflection. LangPoli achieved the best results, with LangPoli (full) reaching a 69.43\% F1 score (12.1 points improvement over Reflexion). From LangPoli's strong performance, we observe the importance of an integrative design for this task where verbal feedback is transformed into actionable insights through dynamic tool creation.

To understand \emph{where} methods fail, we decompose errors in Appendix~\ref{app:confusion} (Table~\ref{tab:structured_info_confusion}). The dominant failure mode is missed patients: ID-only errors remain below 7\% for every method, whereas missed-patient rates vary dramatically from 17.2\% for LangPoli (full) to 93.8\% for LATM.

\subsection{Performance on public datasets: LitSearch, BEIR, and TREC}
\begin{table}[t]
\centering
\caption{Performance on LitSearch. Au-S means the queries that are author-curated (as opposed to computer-generated) and specific. Avg. S means the average result on all specific queries. Improvement=LangPoli -- RankGPT. LangPoli improves over RankGPT in every setting.}
\label{tab:litsearch_main}
\small
\setlength{\tabcolsep}{4pt}
\resizebox{\columnwidth}{!}{%
\begin{tabular}{ll|cc|cc}
\toprule
& & \multicolumn{2}{c|}{\textbf{GPT-4o-mini}} & \multicolumn{2}{c}{\textbf{GPT-4o}} \\
\textbf{Retrieval} & \textbf{Reranking} & \textbf{Au-S} & \textbf{Avg. S} & \textbf{Au-S} & \textbf{Avg. S} \\
 &  & \textbf{R@5} & \textbf{R@5} & \textbf{R@5} & \textbf{R@5} \\
\midrule
BM25 & None & 62.6 & 50.0 & 62.6 & 50.0 \\
BM25 & RankGPT & 71.6 & 65.3 & 76.8 & 68.0 \\
BM25 & \textbf{LangPoli} & \textbf{76.8} & \textbf{67.5} & \textbf{82.9} & \textbf{73.0} \\
 & \textit{Improvement} & \textit{+5.2} & \textit{+2.2} & \textit{+6.1} & \textit{+5.0} \\
\midrule
GritLM & None & 82.5 & 74.8 & 82.5 & 74.8 \\
GritLM & RankGPT & 81.0 & 75.2 & 85.8 & 79.2 \\
GritLM & \textbf{LangPoli} & \textbf{82.0} & \textbf{76.6} & \textbf{88.2} & \textbf{82.0} \\
 & \textit{Improvement} & \textit{+1.0} & \textit{+1.4} & \textit{+2.4} & \textit{+2.8} \\
\bottomrule
\end{tabular}%
}
\end{table}

To show the generality of the proposed method on literature curation, we benchmark LangPoli on the re-ranking task of three publicly available datasets: LitSearch, BEIR, and TREC (full results in Appendix~\ref{app:public}). LangPoli outperforms the strong RankGPT baseline on LitSearch across both GPT-4o-mini and GPT-4o (Table~\ref{tab:litsearch_main}), consistently matches or outperforms RankGPT on most dataset–model combinations of the three BEIR subsets with the longest document contexts. On TREC DL19, LangPoli surpasses RankGPT with GPT-4o despite relying on a base model (GPT-4o-mini) that is approximately 20 times cheaper (Table \ref{tab:dl19_dl20}). Overall, these results show that LangPoli generalizes beyond life science domain. 

\subsection{Tool Quality and Computational Cost}
All created tools maintained perfect execution success and scored in the "green" zone (MI $\geq$ 20) of the \textit{Maintainability Index} (Appendix~\ref{app:tool_quality}). LangPoli-Basic is actually \emph{cheaper} per iteration than ReAct (16{,}653 vs.\ 24{,}318 tokens) while achieving higher F1, because actors receive focused guidelines rather than re-ingesting the full manuscript each round, and LangPoli-Full uses more tokens (55{,}516) in exchange for F1 gains (Appendix~\ref{app:cost}). Appendix~\ref{app:trajectory} contrasts the tool usage trajectories of ReAct and LangPoli on a representative paper. 

\subsection{Variants of LangPoli Show Robust Performance}
Appendix~\ref{app:base_llm} (Table \ref{tab:basis_llm_comparison}) shows an ablation study on the base LLM models. Three plain LLMs from Anthropic were used for the analysis. Claude-3, Claude-3.5, and Claude-3.7 showed 24.93\%, 58.22\%, and 77.17\% F1 scores, illustrating the difficulty of the given task. LangPoli substantially improved performance in all cases, bringing F1 score gains of 10.36, 24.94, and 9.21 points, respectively. 

\begin{table}[htbp]
  \centering
  \caption{LangPoli variants analysis on Structured Information Retrieval. Removing MoCA causes the largest F1 drop.}
  \label{tab:variants}
  \small
  \resizebox{\columnwidth}{!}{%
\begin{tabular}{l|c|cccc}
    \toprule
    Method & F1 (\%) & Pat. ID & Resp. & Exact & Missed \\
     & & (\%) & Rate (\%) & Matches & Pat. \\
    \midrule
    full $-$ MoCA                & 62.97 & 74.77 & 82.57 & 10.31 & 3.81 \\
    full $-$ init. explore      & 65.83 & 74.70 & 84.73 & 11.24 & 2.53 \\
    full $-$ late task guide    & 67.57 & \textbf{84.27} & \textbf{90.40} & \textbf{11.67} & \textbf{1.70} \\
    full $-$ detailed policy    & 66.53 & 76.77 & 86.30 & 10.96 & 2.86 \\
    \midrule
    full $-$ feedback $+$ refl. & \textbf{71.33} & 81.77 & \textbf{91.20} & 11.60 & 2.18 \\
    full $+$ reflection         & 66.43 & 74.80 & 80.63 & 11.50 & 2.24 \\
    \midrule
    \textbf{LangPoli (full)} & \textbf{69.43} & \textbf{81.77} & 90.30 & \textbf{12.03} & \textbf{1.63} \\
    \bottomrule
  \end{tabular}%
}
\end{table}

Table~\ref{tab:variants} presents a performance analysis of LangPoli variants. Using a single actor (LangPoli $-$ MoCA) decreases the F1 score by 6.46 points, highlighting the benefit of specialized parallel actors. Removing the initial exploration phase (LangPoli $-$ init. explore) reduces F1 by 3.60 points. Placing task guidelines early rather than late in the prompt (LangPoli $-$ late task guide) decreases performance by 1.86 points, a phenomenon also known as \textit{recency bias}~\cite{recency_bias}. 
Simplified policy prompts (LangPoli $-$ detailed policy) reduce performance by 2.90 points, confirming the importance of prompt engineering for LLM reasoning. Simply adding self-reflection as in Reflexion (LangPoli $+$ reflection) decreases performance due to \textit{feedback interference}, where policy agent self-reflection conflicts with critic feedback. However, replacing critic feedback with self-reflection (LangPoli $-$ feedback $+$ refl.) yields a slight improvement, suggesting opportunities for task-specific optimization. This ablation study highlights the agentic components and designs that help the task of scientific data curation.

\subsection{Curriculum Transfer Improves Performance on Related Tasks}
After solving a source task $G_{\text{src}}$ on a set of environments, LangPoli accumulates tool creation sequences containing verbal actions, generated code, and observations from all exploration episodes. Curriculum transfer distills this experience into generalizable strategies, combines them with the target task description into a curriculum, and instantiates the curriculum as an initial set of task-adapted tools for a related target task $G_{\text{tgt}}$ (formalization and full results in Appendix~\ref{app:curriculum}, Table~\ref{tab:curriculum}). We evaluate curriculum transfer from "scRNA-seq + immunotherapy response" to the related but distinct "scRNA-seq + scTCR-seq" classification task. While ReAct improved over the plain LLM (78.46\% vs.\ 73.46\% F1) through tool utilization, LangPoli Curriculum exhibited superior results across all curriculum lengths, reaching 87.12\% accuracy and 87.55\% F1 score, which is 14.09 points improvement over the baseline. Notably, even short curricula (5 iterations) performed competitively, suggesting efficient transfer of procedural knowledge between domains. 

\section{Conclusion}
We introduce SciCurateBench, a challenging benchmark that simulates realistic scientific data curation, which we believe is the missing piece in current scientific AI agents research. SciCurateBench requires precise verification of conjunctive experimental conditions over long, heterogeneous documents (manuscript, supplementary files, and raw data). Our experiments show that this task is difficult, even for advanced LLMs and popular agents, and that the primary failure mode is the inability to identify key evidence buried in supplementary files. To study what closes this gap, we propose LangPoli, a multi-agent framework that adapts established agentic components to scientific data curation by decoupling a natural-language policy from per-document code synthesis. LangPoli shows strong performance on both benchmark tasks, transfers to public re-ranking benchmarks, and, through curriculum transfer, adapts to a related task within a few iterations.

\section*{Limitations}

Several limitations should be noted. First, SciCurateBench currently contains 200 papers. Although its total content volume substantially exceeds that of many existing benchmarks, scale expansion is an important direction for future work. Second, the current benchmark focuses on life science research, and extending it to additional scientific domains remains an important direction. Third, LangPoli depends on capable base models, since it must generate goal-oriented, error-free code. The scope of our work is to address the important yet underexplored task of scientific data curation with a new benchmark dataset and to provide baseline methods. Although each component of LangPoli is based on established agentic AI works, we believe its integrative design yields meaningful insights into the problem. Future work may evaluate more recent methods in agentic programming and RAG.

\section*{Acknowledgments}
This work was supported by a Discovery grant from the Natural Sciences and Engineering Research Council (NSERC) of Canada, and a department startup fund from the University of British Columbia (to J.D.). 
J.D. is a Canada Research Chair and is supported by the Canadian Institutes of Health Research through the Canada Research Chair Program. 
The computational resource is partially supported by the Canada Foundation for Innovation \& John. R. Evans Leader Fund (to J.D.). 
This research was supported in part through the computational resources and services provided by Advanced Research Computing at the University of British Columbia and by the Digital Research Alliance of Canada (alliancecan.ca).  


\bibliography{custom}

\appendix

\section{Extended Related Work}
\label{app:related_work}

\textbf{LLM reasoning.}
Recent advances have enhanced LLM reasoning through structured prompting and tool integration. Chain-of-Thought (CoT) \cite{cot} decomposes complex problems into sequential sub-steps. ReAct \cite{react} pioneered cyclical interleaving of reasoning, action, and observation, while ChatCoT \cite{chatcot} showed that conversational formats improve tool interaction. Feedback-based refinement has further advanced reasoning capabilities: Reflexion \cite{reflexion} employs explicit verbal feedback across multiple cycles, and SelfRefine \cite{self_refine} uses internal correction loops. More recently, Agent-as-a-Judge \cite{agent_as_a_judge} uses agentic systems to evaluate other agentic systems and provide intermediate feedback throughout the task-solving process, and DPSDP \cite{reinforce_llm} trains an actor-critic pair of LLMs with reinforcement learning to iteratively refine answers. We thus note that general verbal feedback is an established concept in the field and point the readers to a survey \cite{feedback_survey} for more related works. Our work studies the effectiveness of verbal info-gap feedback within the scope of scientific data curation.

\textbf{Tool creation.}
A separate line of work explores tool creation. LATM \cite{latm} creates generalized tools for problem classes, though this approach falters when "master tools" are impractical to construct. CodeAct \cite{codeact} generates code as actions instead of relying on the fixed JSON format, thereby revising prior actions or emitting new ones upon new observations. ToolMaker \cite{toolmaker} takes a task description and a link to a code repository, and generates code to turn the repository into an LLM-compatible tool. DynaSaur \cite{dynasaur} lets an agent create new actions as needed by writing and executing programs, and accumulates the generated actions for reuse in later tasks. LangPoli is structurally close to these methods but decouples policy (guidelines) from environment-specific implementation (code).

\textbf{Scientific AI agents.}
Recent works have begun applying LLM agents to scientific workflows. ChemCrow \cite{chemcrow} integrates expert-designed chemistry tools for synthesis planning and execution, including direct interaction with a robotic synthesis platform. The AI Scientist \cite{aiscientist} demonstrates end-to-end automation of machine learning research, from hypothesis generation to paper writing. BixBench \cite{bixbench} is a benchmark dataset comprising 60 real-world bioinformatics analysis scenarios with 200 open-ended queries. ScienceAgentBench \cite{scienceagentbench} evaluates agentic systems on data-driven scientific discovery tasks, such as ``Train a cell counting model on the BBBC002 datasets \cite{bbbc002}" or ``Train a multitask model on the Clintox dataset \cite{clintox} to predict a drug's toxicity and FDA approval status". Existing research on scientific AI agents predominantly targets downstream scientific tasks of hypothesis generation, experimental design, data analysis. It is typically assumed that curated input data is already available. We think the upstream bottleneck of curating experimental data from raw literature remains underexplored. 

\section{Dataset Details}
\label{app:dataset}

\subsection{Annotation Protocol}
\label{app:annotation}
Three domain experts with a life science background (3--12 years experience) performed annotation. The annotation process involved: (1) examining full manuscripts, supplementary materials, and hosted data to verify experimental assay types, (2) determining whether data were newly generated or reanalyzed from prior publications, (3) validating clinical outcome availability, and (4) extracting structured metadata (sample/patient identifiers and response information). Annotations were conducted in two stages: primary annotation (160 hours) followed by independent verification with disagreement resolution (additional 40 hours). The senior annotator's labels were merged into the final consensus through joint discussion. The resulting two annotations yielded a Cohen's $\kappa$ of 0.932. Discrepancies arose primarily from ambiguous data provenance statements and unstandardized naming conventions (e.g., ``sample ID'' vs ``patient ID'' vs ``donor ID''). Approximately 10\% of annotations required discussion and joint re-examination to reach consensus. 

\subsection{Dataset Characteristics}
\label{app:characteristics}
\begin{figure*}[t]
\centering
\includegraphics[width=0.98\textwidth]{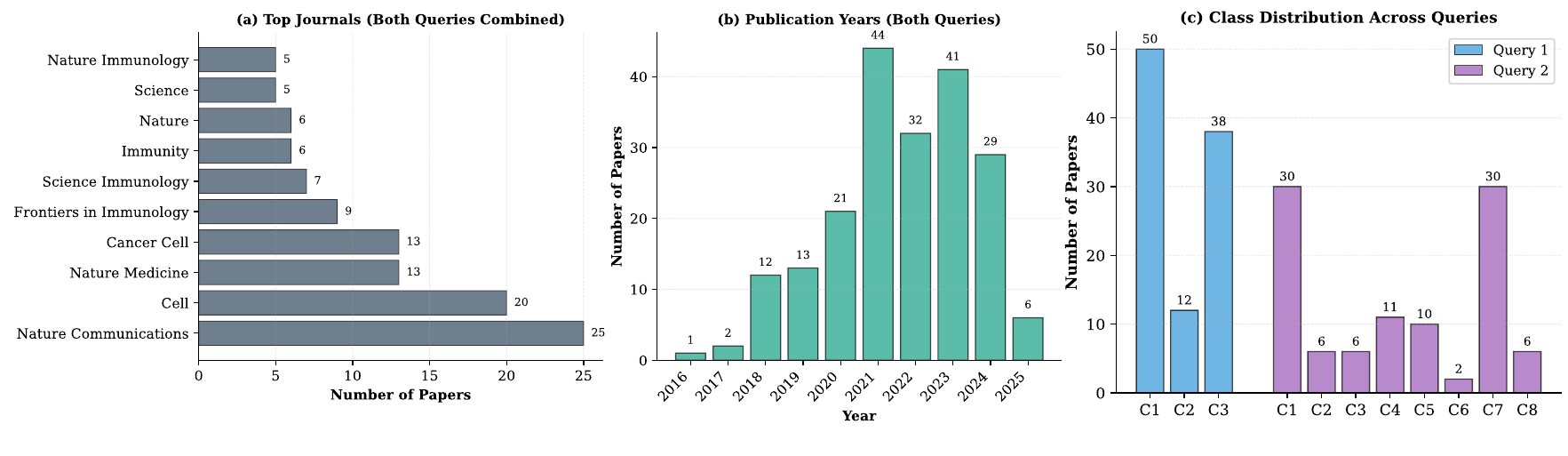}
\caption{SciCurateBench dataset statistics: (a) journal distribution, (b) publication years, (c) class distributions for both queries}
\label{fig:dataset_stats}
\end{figure*}

For Query 2 (paired scRNA-seq + scTCR-seq), papers are distributed across 8 disease categories: healthy controls (Class 2, $n=6$), infections excluding COVID-19 (Class 3, $n=6$), COVID-19 (Class 4, $n=11$), inflammation (Class 5, $n=10$), immune-related adverse events (Class 6, $n=2$), solid tumors (Class 7, $n=30$), other conditions (Class 8, $n=6$), and no qualifying data (Class 1, $n=30$). Summary statistics can be found in Figure~\ref{fig:dataset_stats}. 

\subsection{Dataset Setup}
\label{app:setup}
The download links to all materials for each paper, including manuscripts, supplementary files, and hosted data, are included in the release. These links typically require journal website access permission. To comply with copyright restrictions, users should manually download files through a web browser. The preprocessing script converts the downloaded materials into formats suitable for experimental use. Classification labels are provided in separate CSV files, while Structured Information Retrieval labels are supplied in a PKL file. We convert all PDF files to plain text using the Python \texttt{PyMuPDF} package \cite{pymupdf}. For XLSX files, we separate individual sheets and convert them to text files. We chose text format over CSV because many supplementary tables contain complex structures that are difficult to parse. We also annotated key files containing critical information (e.g., sample/patient IDs and immunotherapy response rates).

\subsection{Why Class 3 Membership Is Hard to Verify}
\label{app:difficulty}
Many papers that look like Class 3 do not publish their full metadata (such as per-sample response labels) for diverse reasons. Some release their assay data but withhold clinical outcomes as proprietary. Others report only aggregate demographics rather than individual-level metadata. In some cases, the patient-level information resides only in the hosted raw data, which is exactly what makes manual curation so time-consuming.

\section{LangPoli Details}
\label{app:langpoli}

\subsection{Example of a Verbal Action}
\label{app:verbal_action}
The following is an example of a verbal action $\tilde{a}_t$ generated by the policy agent:

{\scriptsize
\begin{verbatim}
ã_t = "Create a tool that reads:
The CD4 and CD8 metadata files to understand
their structure:
 - GSE200996_CD4.PBMC.single.cell.meta.data.txt
 - GSE200996_CD8.PBMC.single.cell.meta.data.txt 
 - GSE200996_CD4.tumor.single.cell.meta.data.txt
 - GSE200996_CD8.tumor.single.cell.meta.data.txt

Return:
- Full header information 
- First few rows of data
- Column names and data types
- Any notes or descriptions at the top of files

The goal is to understand how single-cell data
is annotated with patient and response
information."
\end{verbatim}
}

\subsection{Critic Feedback Schema}
\label{app:critic}
Given accumulated evidence $V_{t+1}$ and task objective $G$, the critic produces structured feedback:

\begin{equation}
\resizebox{0.8\columnwidth}{!}{$
\begin{aligned}
r_t = \text{critic}(V_{t+1}, G; \theta_c) = \{\, &
\text{evidence\_summary}: \text{key discoveries}, \\
& \text{info\_gaps}: \{p_i \in G : p_i \text{ not verified}\}, \\
& \text{sufficiency}: \mathbb{I}[\text{sufficient to judge } G], \\
& \text{confidence}: \rho_t \in [0, 1], \\
& \text{next\_steps}: \text{ guidance for } \pi
\,\}
\end{aligned}
$}
\end{equation}

\subsection{Curriculum Transfer}
\label{app:curriculum}
After solving a source task $G_{\text{src}}$ on a set of environments, LangPoli accumulates tool creation sequences $\mathcal{T}_{\text{src}}$ containing verbal actions, generated code, and observations from all exploration episodes. Curriculum transfer leverages this experience to accelerate solving a related target task $G_{\text{tgt}}$:

\textbf{Experience Distillation.} An Experience Analysis Agent $\phi_{\text{exp}}$ processes $\mathcal{T}_{\text{src}}$ to extract generalizable strategies:

\begin{equation}
\resizebox{0.88\columnwidth}{!}{$\mathcal{K}_{\text{src}} = \phi_{\text{exp}}(\mathcal{T}_{\text{src}}) = \{\text{pattern}_1, \text{pattern}_2, \ldots, \text{pattern}_m\}$}
\end{equation}

\noindent where each pattern captures effective exploration heuristics (e.g., ``response metrics often appear in supplementary tables'').

\textbf{Curriculum Generation.} A Curriculum Policy Agent $\phi_{\text{curr}}$ combines distilled knowledge with target task description:

\begin{equation}
\mathcal{U}_{\text{tgt}} = \phi_{\text{curr}}(\mathcal{K}_{\text{src}}, G_{\text{tgt}}, \mathcal{C}_{E}^{\text{tgt}})
\end{equation}

The curriculum $\mathcal{U}_{\text{tgt}}$ is a structured guide specifying: (1) which document sections to prioritize, (2) what patterns to look for, and (3) how to adapt strategies from $G_{\text{src}}$ to $G_{\text{tgt}}$.

\textbf{Tool Implementation.} A Tool Creation Agent $\phi_{\text{tool}}$ instantiates $\mathcal{U}_{\text{tgt}}$ as executable code:

\begin{equation}
c_{\text{init}}^{\text{tgt}} = \phi_{\text{tool}}(\mathcal{U}_{\text{tgt}}, \mathcal{C}_{E}^{\text{tgt}})
\end{equation}

This provides LangPoli with an initial set of task-adapted tools, reducing the number of exploration iterations required for $G_{\text{tgt}}$.


\begin{table}[t]
\centering
\caption{Performance boosting by curriculum transfer for scRNA-seq + scTCR-seq classification}
\label{tab:curriculum}
\footnotesize
\resizebox{\columnwidth}{!}{%
\begin{tabular}{l|cccc}
\toprule
\textbf{Method} & \textbf{Acc (\%)} & \textbf{Prec (\%)} & \textbf{Rec (\%)} & \textbf{F1 (\%)} \\
\midrule
Plain LLM                & 71.28 & 83.17 & 70.01 & 73.46 (1.71) \\
\midrule
ReAct                    & 82.83 & 82.47 & 76.59 & 78.46 (5.28) \\
\midrule
LangPoli Curriculum (5 iter)   & 87.12 & 91.16 & 84.29 & 86.81 (1.68) \\
LangPoli Curriculum (7 iter)   & 85.80 & \textbf{93.21} & 82.98 & 86.95 (1.04) \\
LangPoli Curriculum (10 iter)  & \textbf{87.12} & 90.78 & \textbf{85.64} & \textbf{87.55} (1.26) \\
\bottomrule
\end{tabular}%
}
\end{table}

\section{Experimental Details}
\label{app:exp_details}
The base LLMs are GPT-4o \cite{gpt4_technical_report} [gpt-4o-2024-11-20, OpenAI], Claude-3 [claude-3-haiku-20240307, Anthropic], Claude-3.5 [claude-3-5-sonnet-20241022, Anthropic], and Claude-3.7 \cite{claude3} [claude-3-7-sonnet-20250219, Anthropic]. For BEIR, we also evaluate several smaller LLMs: GPT-4o-mini, Qwen2.5-Coder \cite{qwen2,qwen2_2}, Qwen2.5-7B, Zephyr-7B \cite{zephyr}, Hermes-2-Pro \cite{Hermes2ProLlama3_8B}, OpenChat-3.5-0106 \cite{openchat}, and DeepSeek-Coder-6.7b-Instruct \cite{deepseekcoder}. LitSearch and TREC primarily used GPT-4o-mini and GPT-4o. Experiments are conducted on a server with 64 Intel Xeon Gold 6526Y CPUs and 4 NVIDIA L40S GPUs. 

\section{Results on Public Datasets: LitSearch, BEIR, and TREC}
\label{app:public}

To show the generality of the proposed method on literature curation, we benchmark LangPoli on the re-ranking task of three publicly available datasets: LitSearch, BEIR, and TREC. The re-ranking task asks the system to re-rank the top candidates, retrieved by the base method such as BM25, to determine the final permutation. 

We observe that LangPoli outperforms the strong RankGPT baseline (Table~\ref{tab:litsearch_main}). Improvements are clearer when the underlying retrieval method is simpler (e.g., BM25 compared to GritLM), suggesting that LangPoli effectively compensates for weaker initial retrieval while still providing gains on top of neural retrievers. Table~\ref{tab:litsearch_detailed} shows LangPoli improves performance compared to the baseline BM25 and RankGPT across query sources and paper venues. 

Table~\ref{tab:beir_benchmark} reports re-ranking performance on three BEIR subsets (Touche, News, and Robust04) that have the longest document contexts in the dataset (640.24, 1123.73, and 806.84 average words, respectively). Across a diverse set of base LLMs, LangPoli consistently matches or outperforms RankGPT on most dataset–model combinations. 

Table~\ref{tab:dl19_dl20} shows that LangPoli with GPT-4o-mini consistently outperforms RankGPT using the same base model. 

Overall, these results show that LangPoli generalizes beyond life science domain. More importantly, building on established agentic components, our integrative design benefits literature curation and information retrieval tasks.

\begin{table}[t]
\centering
\caption{Performance breakdown on LitSearch specific, high-quality queries (Specificity=1, Quality=2) by query source and paper venue. GPT-4o-mini was used as the basis LLM. }
\label{tab:litsearch_detailed}
\small
\setlength{\tabcolsep}{3pt}
\resizebox{\columnwidth}{!}{%
\begin{tabular}{ll|cc|cc}
\toprule
& & \multicolumn{2}{c|}{\textbf{GPT-generated}} & \multicolumn{2}{c}{\textbf{Author-curated}} \\
\textbf{Retrieval} & \textbf{Reranking} & \textbf{ACL} & \textbf{Non-ACL} & \textbf{ACL} & \textbf{ICLR} \\
\midrule
BM25 & None & 37.7 & 31.7 & 49.3 & 69.4 \\
BM25 & RankGPT & 58.2 & 53.9 & 58.2 & 72.2 \\
BM25 & \textbf{LangPoli} & \textbf{55.7} & \textbf{55.0} & \textbf{64.2} & \textbf{75.0} \\
 & \textit{Improvement} & \textit{-2.5} & \textit{+1.1} & \textit{+6.0} & \textit{+2.8} \\
\midrule
GritLM & None & 63.1 & 61.7 & 77.6 & 80.6 \\
GritLM & RankGPT & 66.4 & 63.9 & 70.1 & \textbf{91.7} \\
GritLM & \textbf{LangPoli} & \textbf{70.5} & \textbf{66.1} & \textbf{79.1} & 88.9 \\
 & \textit{Improvement} & \textit{+4.1} & \textit{+2.2} & \textit{+9.0} & \textit{-2.8} \\
\bottomrule
\end{tabular}%
}
\end{table}

\begin{table}[htbp]
  \centering
  \caption{Performance (NDCG@10) on BEIR subsets with long corpus contexts. RGP = RankGPT, LP = LangPoli. $\Delta$ denotes LP -- RGP. Bold indicates improvement by LangPoli.}
  \label{tab:beir_benchmark}
  \small
  \setlength{\tabcolsep}{4pt}
  \begin{tabular}{ll|ccc}
    \toprule
    \textbf{Dataset} & \textbf{Base LLM} & RGP & LP & $\Delta$ \\
    \midrule
    \multirow{7}{*}{Touche}
      & GPT-4o-mini    & 34.12 & 37.50 & \textbf{+3.38} \\
      & Qwen2.5-Coder  & 37.88 & 40.54 & \textbf{+2.66} \\
      & Zephyr-7B      & 40.89 & 66.11 & \textbf{+25.22} \\
      & Hermes-2-Pro   & 44.30 & 39.87 & -4.43 \\
      & OpenChat-3.5   & 44.06 & 41.11 & -2.95 \\
      & DeepSeek-Coder & 28.13 & 33.05 & \textbf{+4.92} \\
      & Qwen2.5-7B     & 29.04 & 35.42 & \textbf{+6.38} \\
    \midrule
    \multirow{7}{*}{News}
      & GPT-4o-mini    & 48.80 & 49.03 & \textbf{+0.23} \\
      & Qwen2.5-Coder  & 44.60 & 47.67 & \textbf{+3.07} \\
      & Zephyr-7B      & 34.47 & 48.94 & \textbf{+14.47} \\
      & Hermes-2-Pro   & 40.38 & 44.36 & \textbf{+3.98} \\
      & OpenChat-3.5   & 38.51 & 42.44 & \textbf{+3.93} \\
      & DeepSeek-Coder & 33.31 & 36.88 & \textbf{+3.57} \\
      & Qwen2.5-7B     & 44.56 & 46.70 & \textbf{+2.14} \\
    \midrule
    \multirow{7}{*}{Robust04}
      & GPT-4o-mini    & 57.11 & 57.26 & \textbf{+0.15} \\
      & Qwen2.5-Coder  & 48.48 & 48.56 & \textbf{+0.08} \\
      & Zephyr-7B      & 39.42 & 43.02 & \textbf{+3.60} \\
      & Hermes-2-Pro   & 45.49 & 49.65 & \textbf{+4.16} \\
      & OpenChat-3.5   & 43.61 & 43.74 & \textbf{+0.13} \\
      & DeepSeek-Coder & 42.41 & 41.69 & -0.72 \\
      & Qwen2.5-7B     & 48.77 & 50.37 & \textbf{+1.60} \\
    \bottomrule
  \end{tabular}
\end{table}

\begin{table}[t]
\centering
\caption{Performance comparison on TREC-DL benchmarks. NDCG@10 is used as the metric. $\Delta$ denotes improvement of LangPoli (gpt-4o-mini) over RankGPT (gpt-4o-mini).}
\label{tab:dl19_dl20}
\small
\setlength{\tabcolsep}{6pt}
\begin{tabular}{lcc}
\toprule
\textbf{Method} & \textbf{DL19} & \textbf{DL20} \\
\midrule
BM25 & 50.58 & 47.96 \\
RankGPT (gpt-3.5-turbo) & 65.80 & 62.91 \\
RankGPT (gpt-4o-mini) & 72.61 & 67.59 \\
RankGPT (gpt-4o) & 75.59 & 70.56 \\
\midrule
LangPoli (gpt-4o-mini) & \textbf{75.93} & \textbf{68.79} \\
\textit{$\Delta$ (vs. RankGPT gpt-4o-mini)} & \textbf{+3.32} & \textbf{+1.20} \\
\bottomrule
\end{tabular}
\end{table}

\section{Additional Analyses}
\label{app:analyses}

\subsection{Error Breakdown on Structured Information Retrieval}
\label{app:confusion}
Table~\ref{tab:structured_info_confusion} decomposes errors into three categories conditioned on ground-truth patients -- correct (both ID and response right), ID-only (correct patient retrieved but response misclassified), and missed (patient not retrieved at all) -- together with false-positive rate (FP) over predicted patients. A consistent pattern emerges across all methods. The dominant failure mode is missed patients. ID-only errors remain below 7\% for every method, indicating that once a patient is correctly retrieved, the associated response rate is usually extracted correctly. In contrast, missed-patient rates vary dramatically from 17.2\% for LangPoli (full) to 93.8\% for LATM. 

\begin{table}[htbp]
  \centering
  \caption{Confusion breakdown on Structured Information Retrieval. Correct, ID-only, and Missed are computed over ground-truth patients and sum to 100\%. FP is computed over predicted patients. }
  \label{tab:structured_info_confusion}
  \small
  \resizebox{\columnwidth}{!}{%
\begin{tabular}{l|ccc|c}
    \toprule
    \textbf{Method} & \textbf{Correct} & \textbf{ID-only} & \textbf{Missed} & \textbf{FP} \\
                    & \textbf{(\%)}    & \textbf{(\%)}    & \textbf{(\%) $\downarrow$} & \textbf{(\%) $\downarrow$} \\
    \midrule
    ReAct           & 49.9 & 6.8 & 43.3 & 26.1 \\
    ChatCoT         & 50.8 & 5.0 & 44.2 & 32.5 \\
    LATM            &  4.3 & 1.9 & 93.8 & 60.7 \\
    \midrule
    Reflexion       & 57.1 & 4.1 & 38.8 & 19.0 \\
    SelfRefine      & 41.7 & 4.3 & 54.0 & 28.9 \\
    \midrule
    LangPoli-Basic  & 70.3 & 3.8 & 25.9 & 32.6 \\
    LangPoli (full) & \textbf{76.6} & 6.2 & \textbf{17.2} & 28.2 \\
    \bottomrule
  \end{tabular}%
}
\end{table}

\subsection{Base LLM Ablation}
\label{app:base_llm}
Table~\ref{tab:basis_llm_comparison} reports the performance of plain LLMs and LangPoli with three base LLMs from Anthropic. 

\begin{table}[t]
\centering
\caption{Performance of plain LLM and LangPoli by different base LLM models on the SciCurateBench dataset}
\label{tab:basis_llm_comparison}
\footnotesize  
\resizebox{\columnwidth}{!}{%
\begin{tabular}{l l c c c c}
\toprule
\textbf{Model} & \textbf{Method} & \textbf{Acc (\%)} & \textbf{F1 (\%)} & \textbf{C3 F1 (\%)} & \textbf{Gain (\%)} \\
\midrule
\multirow{2}{*}{Claude-3} 
& Plain & 25.33 & 24.93 (1.40) & 38.24 & \multirow{2}{*}{+10.36} \\
& LangPoli    & \textbf{37.67} & \textbf{35.29} (2.32) & \textbf{45.27} & \\
\midrule
\multirow{2}{*}{Claude-3.5} 
& Plain & 67.67 & 58.22 (2.23) & 66.59 & \multirow{2}{*}{+24.94} \\
& LangPoli    & \textbf{86.33} & \textbf{83.16} (0.94) & \textbf{88.19} & \\
\midrule
\multirow{2}{*}{Claude-3.7} 
& Plain & 80.00 & 77.17 (2.22) & 79.51 & \multirow{2}{*}{+9.21} \\
& LangPoli    & \textbf{87.29} & \textbf{86.38} (1.00) & \textbf{87.65} & \\
\bottomrule
\end{tabular}%
}
\end{table}

\subsection{Tool Quality}
\label{app:tool_quality}
Table \ref{tab:tool_creation_t3} summarizes the scores of created tools. We can see from high Rule Compliance values that each agent successfully interpreted its role-specific guidance. 
Notably, all implementations maintained perfect execution success, showing the validity of our safety guideline. 
We also measured the \textit{Maintainability Index} (MI), which measures how maintainable the source code is. Although MI is not fully optimized for Python's dynamic nature, the created tools scored remarkably high. The tools produced by all actors scored in the "green" zone (MI $\geq$ 20), indicating readability, testability, and modifiability. 

\begin{table}[htbp]
  \centering
  \caption{Evaluation of tool creation in Structured Information Retrieval. Actor 1, Actor 2, and Actor 3 are Manuscript Explorer, Supplementary Content Reader, and Large File Summarizer.}
  \label{tab:tool_creation_t3}
  \small
  \setlength{\tabcolsep}{4pt}  
  \resizebox{\columnwidth}{!}{%
\begin{tabular}{l|cc|cccc}
    \toprule
    \textbf{Agent} & \textbf{Exec.} & \textbf{Evid.} & \textbf{Cyclo.} & \textbf{Hal.} & \textbf{LoC} & \textbf{MI} \\
     & \textbf{Succ. (\%)} & \textbf{Len.} & \textbf{Complex.} & \textbf{Vol.} & & \\
    \midrule
    Actor 1 & 100.00 & 45,780 & 7.36 & 543.53 & 46.77 & 43.71 \\
    Actor 2 & 100.00 & 78,271 & 4.22 & 385.67 & 38.17 & 47.38 \\
    Actor 3 & 100.00 & 35,871 & 6.72 & 533.25 & 49.16 & 43.46 \\
    LangPoli (full) & 100.00 & 53,320 & 6.1 & 397.57 & 44.70 & 48.28 \\
    \bottomrule
  \end{tabular}%
}
\end{table}

\subsection{Computational Cost}
\label{app:cost}
Table~\ref{tab:cost_analysis} reports token consumption alongside classification F1 on SciCurateBench. Each LangPoli round involves five API calls: the Policy Agent generates verbal actions ($\sim$3{,}500 tokens), Actor Agents create tools to execute them ($\sim$3{,}500 tokens each), and the Critic Agent evaluates the collected evidence (variable, depending on evidence size). LangPoli-Basic is actually \emph{cheaper} per iteration than ReAct (16{,}653 vs.\ 24{,}318 tokens) while achieving higher F1. LangPoli-Full uses more tokens (55{,}516) in exchange for F1 gains. We view this trade-off as favorable for scientific data curation. Each additional correctly identified paper contributes substantial patient-level data to downstream atlas-building efforts, which justifies the additional token cost. Token cost itself can be affected by other factors such as the API cost of the base LLMs. 

\begin{table}[t]
\centering
\small
\caption{Token consumption on SciCurateBench Paper Classification.
Prompt and Response columns report average tokens per iteration.
Total = Prompt + Response. }
\label{tab:cost_analysis}
\resizebox{\columnwidth}{!}{%
\begin{tabular}{l|ccc|cc}
\toprule
\textbf{Method} & \textbf{Avg.\ Prompt} & \textbf{Avg.\ Response} & \textbf{Total} & \textbf{F1 (\%)} \\
\midrule
ReAct                & 24{,}318 & --      & 24{,}318 & 76.95      \\
\midrule
LangPoli-Basic       & 14{,}491 & 2{,}162 & 16{,}653 & 79.76      \\
LangPoli (full)      & 51{,}145 & 4{,}371 & 55{,}516 & \textbf{83.16}\\
\midrule
\quad Actor 1         & 15{,}788 & 640     & 16{,}428 & -- \\
\quad Actor 2         & 3{,}583  & 1{,}081 & 4{,}664  & -- \\
\quad Actor 3         & 24{,}608 & 488     & 25{,}096 & -- \\
\bottomrule
\end{tabular}%
}
\end{table}

\subsection{Tool Creation Trajectory}
\label{app:trajectory}
\begin{figure*}[t]
    \centering
    \includegraphics[width=0.98\textwidth]{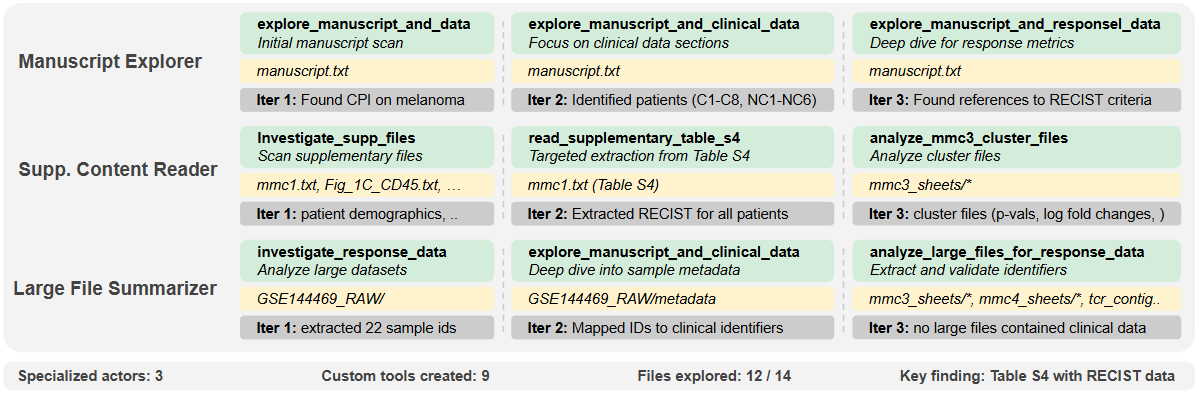}
    \caption{Tool usage trajectories on a representative benchmark paper. }
    \label{fig:trajectory}
\end{figure*}

Figure~\ref{fig:trajectory} shows a trajectory on a representative paper studying checkpoint-inhibitor-induced colitis. LangPoli's three actors follow a coarse-to-fine pattern: Iteration~1 gathered preliminary assay evidence in parallel, Iteration~2 narrowed to data generation context with targeted extraction from Table S4, and Iteration~3 confirmed RECIST responses for patients C1--C8. Across the trajectory, the actors created nine custom tools and explored 12 of 14 available files. Dynamic tool creation surfaces evidence buried in supplementary files that fixed-action agents tend to skip.

\section{Task Guidelines}
\subsection{Paper Classification}
\begin{tcolorbox}[colback=gray!5,colframe=black!20,title=\textbf{Prompt: scRNA-seq + Immunotherapy Classification},label=box:classification_guidance,breakable]
You are given an academic paper (\texttt{paper\_id=\%s}). Your task is to classify it into one of the following three categories:

\begin{itemize}
  \item \textbf{3}: The paper presents newly generated single-cell RNA sequencing (scRNA-seq) data from cancer patients \textbf{before or after immunotherapy}, with each scRNA-seq sample having an associated patient response metric.
  \item \textbf{2}: The paper analyzes existing scRNA-seq data from cancer patients \textbf{before or after immunotherapy}, with response metrics available, but \textbf{the data was not newly generated}.
  \item \textbf{1}: The paper does \textbf{not} analyze scRNA-seq data from cancer patients with associated immunotherapy response information.
\end{itemize}

\textbf{Important notes:}
\begin{itemize}
  \item Assign \textbf{Class 3} if any newly generated scRNA-seq sample has an associated response metric.
  \item Assign \textbf{Class 2} if using previously published scRNA-seq datasets with directly associated response metrics.
  \item Assign \textbf{Class 1} for all other cases.
\end{itemize}

Acceptable response metrics include:
\begin{itemize}
  \item Standardized clinical criteria (e.g., RECIST)
  \item Radiological assessment (e.g., partial response)
  \item Pathological response (e.g., major pathological response)
  \item Explicit labels (e.g., responder vs non-responder)
\end{itemize}

\textbf{Do not} count inferred or indirect markers (e.g., changes in gene expression) unless clinically assessed outcomes are clearly reported.

\vspace{0.3em}
\noindent\textbf{Final Output Format:}
\begin{verbatim}
DOCUMENT CLASS = your_pred (3, 2, or 1)
\end{verbatim}
\end{tcolorbox}

\subsection{Structured Information Retrieval}
\begin{tcolorbox}[colback=gray!5,colframe=black!20,title=\textbf{Prompt: Single-cell RNA + TCR Sequencing Classification},label=box:tcr_guideline,breakable]
You are given an academic paper (\texttt{paper\_id=\%s}). Your task is to classify it into one of the following eight categories:
\begin{itemize}
  \item \textbf{Class 1}: The paper does \textbf{NOT} present any newly generated paired single-cell RNA and TCR sequencing data.
  
  \item \textbf{Class 2}: The paper presents newly generated paired single-cell RNA and TCR sequencing data on healthy people. 
  
  \item \textbf{Class 3}: The paper presents newly generated paired single-cell RNA and TCR sequencing data on patients with infection except COVID-19. 
  
  \item \textbf{Class 4}: The paper presents newly generated paired single-cell RNA and TCR sequencing data on patients with COVID-19 infection. 
  
  \item \textbf{Class 5}: The paper presents newly generated paired single-cell RNA and TCR sequencing data on patients with inflammation. 
  
  \item \textbf{Class 6}: The paper presents newly generated paired single-cell RNA and TCR sequencing data on patients with immune-related Adverse Effect (irAE). 
  
  \item \textbf{Class 7}: The paper presents newly generated paired single-cell RNA and TCR sequencing data on patients with solid tumor. 
  
  \item \textbf{Class 8}: The paper presents newly generated paired single-cell RNA and TCR sequencing data on patients with other disease conditions (ex. myeloma, leukemia, frailty, organoid, etc). 
\end{itemize}

\textbf{Important notes and clarifications:}
\begin{itemize}
  \item To qualify for \textbf{Class 2, 3, 4, 5, 6, 7, 8}, the paper must present T cells where both transcriptomic data and TCR sequences are available at single-cell resolution from the same cells. The RNA and TCR data must be paired (i.e., both data types come from the same individual cells).
  
  \item Separate datasets where scRNA-seq was performed on one cell population and scTCR-seq on another do \textbf{NOT} qualify for Class 3, as this prevents linking transcriptomic profiles to specific TCR sequences.
  
  \item Bulk RNA-seq or bulk TCR-seq data do \textbf{NOT} qualify for Class 2, 3, 4, 5, 6, 7, 8.
  
  \item For papers using multiple datasets: If \textbf{ANY} portion of the data meets the paired single-cell RNA+TCR criteria and is newly generated, classify as one of Class 2, 3, 4, 5, 6, 7, 8. (The goal is to identify all possible sources of paired data for database construction.)
  
  \item If a paper includes paired single-cell RNA and TCR sequencing data but \textbf{ALL} such data comes from previously published studies (i.e., no newly generated paired data), classify as Class 1.
  
  \item Any paper that does not meet the criteria for Class 2, 3, 4, 5, 6, 7, or 8 should be classified as Class 1.
\end{itemize}

\vspace{0.3em}
\noindent\textbf{Final Output Format:}
\begin{verbatim}
DOCUMENT CLASS = your_pred (1, 2, 3, 
4, 5, 6, 7, or 8)
\end{verbatim}
\end{tcolorbox}

\end{document}